\documentclass{article}
\usepackage{lineno}
\usepackage[utf8]{inputenc}
\usepackage[T1]{fontenc}
\usepackage[english]{babel}  % 设置为英文语言环境
\usepackage{url}
\usepackage{amsmath}
\usepackage{geometry}
\usepackage{graphicx}
\usepackage{multicol}
\usepackage{caption}
\usepackage{times}
\usepackage{float}
\usepackage{subcaption}
\usepackage{rotating}
\usepackage{authblk} % 确保在导言区加载该宏包

\begin{document}  
% \linenumbers % 启用行号
\title{Multi-Contrast Fusion Module: An attention mechanism integrating multi-contrast features for fetal torso plane classification}  

\author{}
\date{}

\author[1]{Shengjun Zhu\thanks{19359446568@163.com}}
\author[3]{Siyu Liu}
\author[2]{Runqing Xiong}
\author[2]{Liping Zheng}
\author[2]{Duo Ma}
\author[3]{Rongshang Chen}
\author[1]{Jiaxin Cai\thanks{caijiaxin@xmut.edu.cn}}
% \author[1]{Shengjun Zhu\thanks{19359446568@163.com}}
% \author[3]{Siyu Liu\thanks{1114595740@qq.com}}
% \author[2]{Runqing Xiong\thanks{2571186304@qq.com}}
% \author[2]{Liping Zheng\thanks{787886024@qq.com}}
% \author[2]{Duo Ma\thanks{ligangduoduo@163.com}}
% \author[3]{Rongshang Chen\thanks{2005120705@xmut.edu.cn}}
% \author[1]{Jiaxin Cai\thanks{caijiaxin@xmut.edu.cn}}

\affil[1]{School of Mathematics and Statistics, Xiamen University of Technology, Xiamen, P.R. China}
\affil[2]{Department of Ultrasound Imaging, The Second Affiliated Hospital of Xiamen Medical College, Xiamen, P.R. China}
\affil[3]{School of Computer and Information Engineering, Xiamen University of Technology, Xiamen, P.R. China}

% \date{This work is supported by the Natural Science Foundation of Fujian Province (2023J05083, 2022J011396, 2023J011434) and the Postgraduate Science and Technology Innovation Program Project of Xiamen University of Technology (YKJCX2023055). \\ Shengjun Zhu, Siyu Liu, and Runqing Xiong are the co-first authors. Corresponding authors: Jiaxin Cai.}

% 致谢和资助信息可放在 \thanks 或专门的 Acknowledgment 环节

\maketitle

\section{Abstract}  
\begingroup
\setlength{\parindent}{0pt}
% Prenatal ultrasound plays a critical role in evaluating fetal structural development and detecting abnormalities. Its use has been shown to significantly reduce perinatal complications and improve neonatal survival rates. Accurate identification of standard fetal torso planes is essential for reliable assessment of fetal development and the delivery of personalized prenatal care. However, due to inherent limitations of ultrasound imaging, such as low contrast and indistinct texture details, extracting fine-grained information remains challenging.
% To address this issue, we propose a novel Multi-Contrast Fusion Module (MCFM) designed to enhance the model's ability to capture and model detailed features. Our module is applied exclusively to the lower layers of the network, operating directly on raw ultrasound images. It assigns attention weights to image representations under varying contrast conditions. Importantly, this design ensures minimal additional parameter overhead.
% We validate our approach on a curated dataset of fetal torso planes. Experimental results show that while the integration of MCFM introduces negligible increases in model complexity, it yields substantial improvements in performance. The proposed method supports clinicians in more accurately identifying fetal torso planes, thereby providing robust technical support for prenatal screening and demonstrating strong potential for clinical application.

\textbf{Purpose:}
Prenatal ultrasound is a key tool in evaluating fetal structural development and detecting abnormalities, contributing to reduced perinatal complications and improved neonatal survival. Accurate identification of standard fetal torso planes is essential for reliable assessment and personalized prenatal care. However, limitations such as low contrast and unclear texture details in ultrasound imaging pose significant challenges for fine-grained anatomical recognition.

\textbf{Methods:}
We propose a novel Multi-Contrast Fusion Module (MCFM) to enhance the model's ability to extract detailed information from ultrasound images. MCFM operates exclusively on the lower layers of the neural network, directly processing raw ultrasound data. By assigning attention weights to image representations under different contrast conditions, the module enhances feature modeling while explicitly maintaining minimal parameter overhead.

\textbf{Results:}
The proposed MCFM was evaluated on a curated dataset of fetal torso plane ultrasound images. Experimental results demonstrate that MCFM substantially improves recognition performance, with a minimal increase in model complexity. The integration of multi-contrast attention enables the model to better capture subtle anatomical structures, contributing to higher classification accuracy and clinical reliability.

\textbf{Conclusions:}
Our method provides an effective solution for improving fetal torso plane recognition in ultrasound imaging. By enhancing feature representation through multi-contrast fusion, the proposed approach supports clinicians in achieving more accurate and consistent diagnoses, demonstrating strong potential for clinical adoption in prenatal screening.
The codes are available at \url{https://github.com/sysll/MCFM}.
\endgroup

\section{Introduction}
Pregnancy prenatal ultrasound plays a vital role in the early detection of fetal growth abnormalities. It provides substantial clinical value by reducing perinatal complications and neonatal mortality \cite{salomon2022isuog}. 
Accurate identification of fetal developmental anomalies supports personalized prenatal management. This approach helps minimize long-term consequences caused by structural or functional abnormalities \cite{dawood2022imaging}.
During routine prenatal ultrasound examinations, standard anatomical planes offer critical diagnostic insights into the development of major fetal organs and tissues \cite{fiorentino2023review}. 
For example, transverse and sagittal views of the kidneys enable evaluation of size, morphology, and renal pelvis width. 
These parameters aid in detecting common urogenital anomalies, including hydronephrosis, polycystic kidney disease, and unilateral renal agenesis \cite{brennan2017evaluation}. Additionally, cortical echogenicity may indirectly indicate renal function \cite{seyedzadeh2020relationship}. 
Evaluation of the spinal plane is essential for identifying neural tube defects, such as spina bifida and myelomeningocele \cite{aboughalia2021multimodality}. Key diagnostic markers include spinal continuity, symmetry, and the integrity of the overlying skin. 
Moreover, fetal abdominal circumference reflects overall growth and is strongly associated with high-risk conditions such as fetal growth restriction, macrosomia, chromosomal abnormalities, and gestational diabetes \cite{griffin2024comparing}.
Accurate extraction of these image features directly influences clinical risk assessment and management planning.

Although ultrasound imaging is widely recognized as the primary tool for fetal development assessment due to its non-invasive, safe, and real-time nature \cite{buijtendijk2024diagnostic, recker2021point}, automated analysis remains challenging. 
Fetal ultrasound images often exhibit low contrast, unclear structural boundaries, and high noise levels. 
These characteristics make the identification and classification of key anatomical structures heavily dependent on experienced sonographers. 
This dependence increases diagnostic workload and introduces notable subjectivity and inconsistency.
Deep learning techniques have been widely adopted to improve analytical efficiency and diagnostic accuracy in medical imaging\cite{bansal2024decoding, madival2025naturepred}. These methods have shown notable success in analyzing lung CT scans \cite{wang2022deep, zhang2019pathogenic}, fundus images \cite{abramowicz2022ocular}, and breast ultrasound \cite{tsunoda2024beyond, drukker2020introduction, zhu2024covidllmrobustlargelanguage}. 
Fetal ultrasound images often suffer from inconsistent quality and low local contrast, making it difficult for models to accurately identify key anatomical details. This limitation can significantly impact classification performance. 
In our investigation, we found that channel attention mechanisms hold promise in addressing such issues by selectively emphasizing important feature representations.
Attention mechanisms are a key technique in modern deep neural networks, offering a novel way to enhance the model’s ability to capture local information \cite{vaswani2017attention}. 
Since SENet introduced the channel attention mechanism \cite{hu2018squeeze}, various modules—such as CBAM \cite{woo2018cbam} and BAM \cite{park2018bam}—have been proposed, significantly improving performance in image recognition and object detection \cite{hu2018gather}\cite{zhang2020relation}\cite{lim2021small}. However, these methods are primarily designed for natural images and exhibit limited adaptability to ultrasound images, which often feature complex structures and low signal-to-noise ratios. 
This limitation is especially pronounced in prenatal ultrasound, where attention mechanisms tailored for such data remain underexplored. 
Although some studies have applied attention mechanisms to fetal brain and cardiac image analysis \cite{bhalla2021automatic}\cite{he2024fetal}\cite{zhang2023improving}, limited research has addressed fetal trunk structure recognition, and existing work remains exploratory \cite{xiao2023application}\cite{chen2021artificial}.

In our study, we found that adjusting the contrast of fetal ultrasound images can significantly enhance the visibility of subtle anatomical details. Based on this observation, we generated multiple contrast-enhanced copies of each ultrasound image using a predefined set of contrast parameters. These image variants were then processed using a channel attention mechanism to assign adaptive importance weights.
It is worth noting that conventional channel attention methods—such as the SE module—are typically applied after feature extraction, operating on intermediate feature maps. In contrast, our attention mechanism is integrated into the lower layers of the network and directly processes raw images prior to feature extraction. At this stage, the input retains a large spatial resolution (e.g., 224 × 224) and contains rich, uncompressed information alongside significant redundancy. Applying Global Average Pooling (GAP) at this point to compress the entire image into a single scalar would result in substantial information loss and fail to capture the nuanced variations across contrast-enhanced inputs. For instance, a 224 × 224 image contains 50,176 pixels, and representing such high-dimensional data with a single value would be insufficient to preserve its essential characteristics.

To address this, we propose the Multi-Contrast Fusion Module (MCFM). This module assigns a dedicated CNN-based meta-learner to each contrast-enhanced image variant, mapping them into multiple feature maps. We then introduce a Global Cross-Channel Pooling (GCCP) strategy to compress each group of feature maps into a single representative variable, which is passed through a neural network to learn attention weights. Importantly, these weights are not used to directly combine the contrast-enhanced image variants; instead, they are applied to their corresponding feature maps. This design enhances the attention mechanism’s capacity to model rich spatial structures and preserve fine-grained details at early network stages. As a result, the proposed approach significantly improves the model’s ability to extract critical details—especially in regions with blurred boundaries. Moreover, by assigning attention weights across different contrast levels, our method improves model interpretability, yielding clearer responses in diagnostically relevant areas.
We integrated MCFM into a ResNet backbone and conducted comparative and ablation experiments. The results demonstrate the effectiveness of our approach. Our main contributions are summarized as follows:

\begin{enumerate}
\item We propose a novel MCFM designed to integrate information from multiple contrast levels, thereby enhancing the model's ability to capture fine-grained features. Since the module is embedded only in the early stages of the network, it introduces minimal additional parameters.

\item We introduce the Global Cross-Channel Pooling (GCCP) strategy, which compresses and averages all feature maps extracted from a contrast-enhanced image. This approach enables efficient inter-channel information fusion and facilitates the construction of more robust representations, particularly under varying image quality conditions.

\item We implement the proposed method on our fetal torso ultrasound dataset and conduct comprehensive comparative and ablation experiments. The results demonstrate the effectiveness of our approach in improving both accuracy and robustness.
\end{enumerate}

\section{Materials}
\begin{figure}[htbp]  
    \centering  
    \includegraphics[width=1\linewidth]{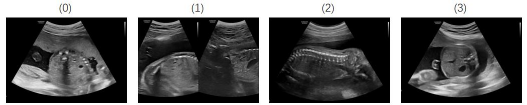} % 假设这个图像包含了两种模型的示意  
    \caption{Representative samples from each class in the dataset. The four classes include: Class 0 – Transverse Kidney, Class 1 – Sagittal Kidney, Class 2 – Sagittal Spine, and Class 3 – Transverse Abdomen. }  
    \label{dataset}  
\end{figure}

We systematically collected 1,200 fetal ultrasound images from pregnant women at the XXXXX between September 2021 and July 2024. 
All procedures strictly adhered to the Declaration of Helsinki and national ethical guidelines for medical research. The dataset comprised clinical images obtained during routine second-trimester level II prenatal ultrasound screenings. 
Images were acquired using high-end color Doppler ultrasound systems (GE Voluson E8 and Samsung WS80A) with convex volume probes (GE: 2-7 MHz; Samsung: 1-8 MHz) and standard convex probes (GE: 2-5 MHz; Samsung: 1-7 MHz).
Inclusion criteria included pregnant women aged 18-50 years with singleton pregnancies at 22-26 weeks’ gestation. 
All participants provided written informed consent. Exclusion criteria included BMI >25 $ \, \text{kg/m}^2$, amniotic fluid depth <2 cm, or amniotic fluid index <5 cm.
The dataset comprised four standard anatomical views: transverse kidney, sagittal kidney, sagittal spine, and transverse abdomen. 
These views were labeled as categories 0-3 for multi-class classification. Images were evenly distributed across categories (n=300 per class). Experienced clinicians annotated all images, with annotation consistency ensured through independent labeling and cross-validation by multiple physicians.
For preprocessing, we resized all ultrasound images to 224$\times$224 pixels to standardize input dimensions. 
Images were then converted to single-channel grayscale format to facilitate subsequent contrast adjustment operations. 
Finally, we normalized pixel values to [0, 1] to improve training stability and accelerate convergence. Example images of these four categories are shown in Figure \ref{dataset}.

\section{Method}  
\subsection{Conception of Contrast}

Image contrast is a fundamental property that describes the difference in luminance or color that makes an object in an image distinguishable from other objects and the background. It plays a critical role in various image processing tasks, such as enhancement, segmentation, and recognition. Higher contrast typically implies better visibility of details and clearer structural information.
A common definition of image contrast, particularly for grayscale images, is based on the standard deviation of pixel intensities. Let $I(x,y)$ denote the intensity of a pixel at location $(x, y)$ in an image with $M \times N$ pixels. The mean intensity $\mu$ of the image is given by
\begin{equation}
\mu = \frac{1}{MN} \sum_{x=1}^{M} \sum_{y=1}^{N} I(x, y).
\end{equation}
Then, the contrast $C$ can be defined as the standard deviation of the image intensities:
\begin{equation}
C = \sqrt{\frac{1}{MN} \sum_{x=1}^{M} \sum_{y=1}^{N} (I(x, y) - \mu)^2}.
\end{equation}

This formulation captures the degree of variation in intensity values, with higher standard deviation indicating greater contrast. In other contexts, especially in local or adaptive contrast enhancement, contrast can also be measured using other metrics such as Michelson contrast or RMS contrast, depending on the application.

\begin{figure}[htbp]
    \centering

    % 第一行：原图 + 标题行
    \begin{subfigure}[b]{0.18\textwidth}
        \centering
        \textbf{Class 0} \\
        \includegraphics[width=\linewidth]{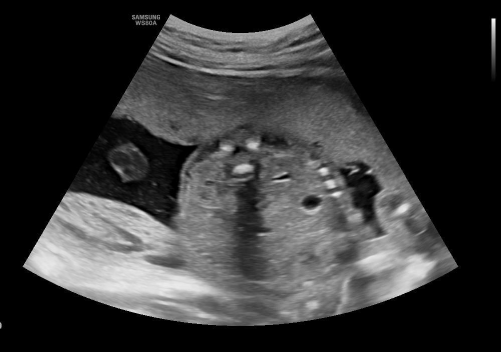}
    \end{subfigure}
    \begin{subfigure}[b]{0.18\textwidth}
        \centering
        \textbf{Class 1} \\
        \includegraphics[width=\linewidth]{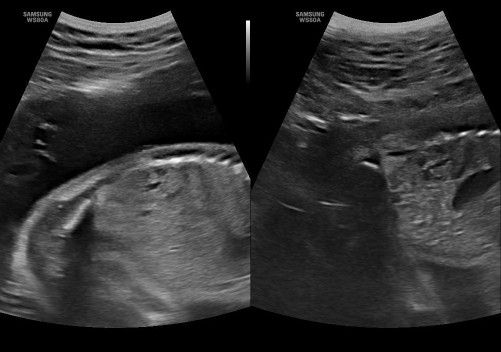}
    \end{subfigure}
    \begin{subfigure}[b]{0.18\textwidth}
        \centering
        \textbf{Class 2} \\
        \includegraphics[width=\linewidth]{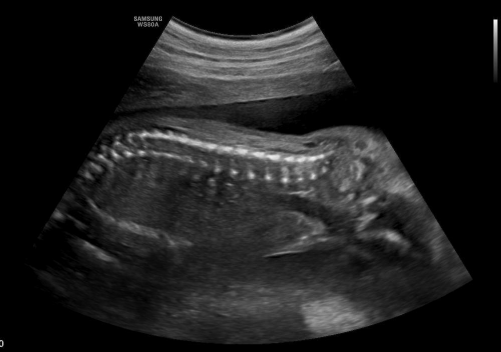}
    \end{subfigure}
    \begin{subfigure}[b]{0.18\textwidth}
        \centering
        \textbf{Class 3} \\
        \includegraphics[width=\linewidth]{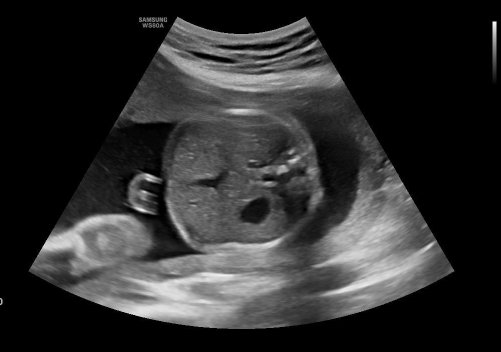}
    \end{subfigure}
        \begin{subfigure}[b]{0.08\textwidth}
        \centering
       {\textbf{Original}}
    \end{subfigure}

    % 第二行：contrast = 1.5
    \begin{subfigure}[b]{0.18\textwidth}
        \centering
        \includegraphics[width=\linewidth]{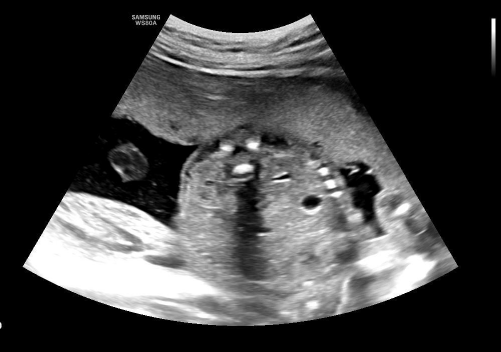}
    \end{subfigure}
    \begin{subfigure}[b]{0.18\textwidth}
        \centering
        \includegraphics[width=\linewidth]{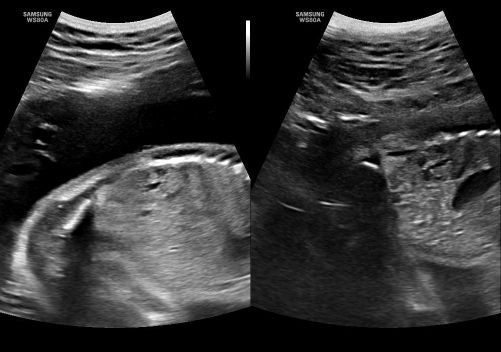}
    \end{subfigure}
    \begin{subfigure}[b]{0.18\textwidth}
        \centering
        \includegraphics[width=\linewidth]{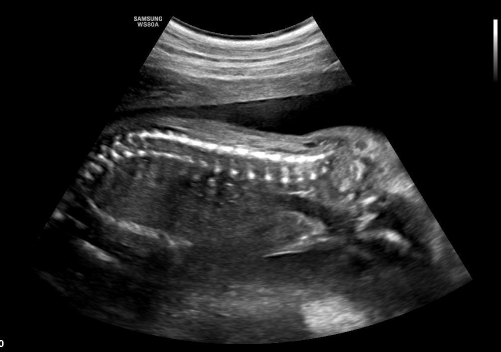}
    \end{subfigure}
    \begin{subfigure}[b]{0.18\textwidth}
        \centering
        \includegraphics[width=\linewidth]{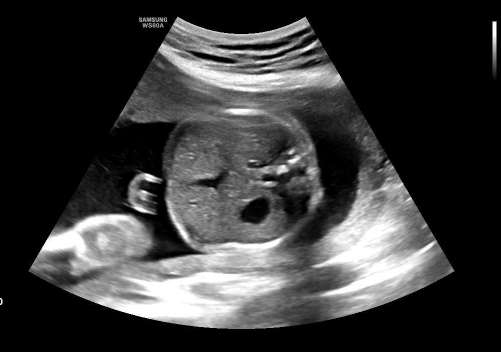}
    \end{subfigure}
    \begin{subfigure}[b]{0.08\textwidth}
        \centering
                {\textbf{C=1.5}}
    \end{subfigure}

    % 第三行：contrast = 2.0
    \begin{subfigure}[b]{0.18\textwidth}
        \centering
        \includegraphics[width=\linewidth]{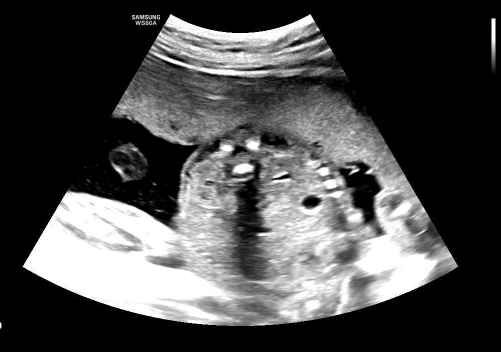}
    \end{subfigure}
    \begin{subfigure}[b]{0.18\textwidth}
        \centering
        \includegraphics[width=\linewidth]{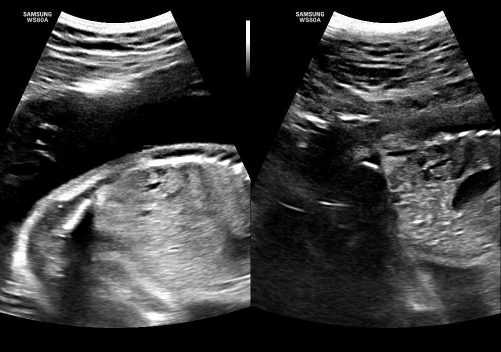}
    \end{subfigure}
    \begin{subfigure}[b]{0.18\textwidth}
        \centering
        \includegraphics[width=\linewidth]{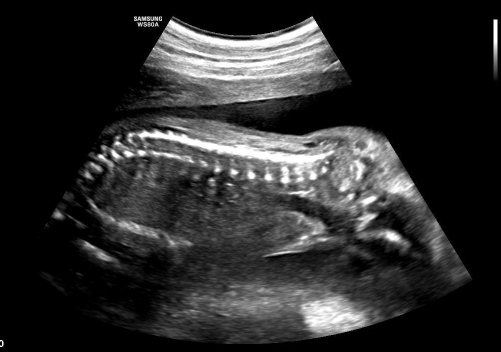}
    \end{subfigure}
    \begin{subfigure}[b]{0.18\textwidth}
        \centering
        \includegraphics[width=\linewidth]{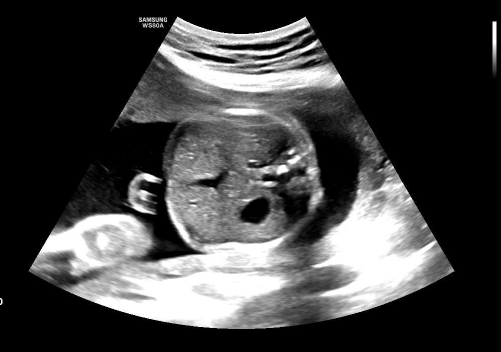}
    \end{subfigure}
    \begin{subfigure}[b]{0.08\textwidth}
        \centering
                {\textbf{C=2.0}}
    \end{subfigure}

    % 第四行：contrast = 2.5
     \begin{subfigure}[b]{0.18\textwidth}
        \centering
        \includegraphics[width=\linewidth]{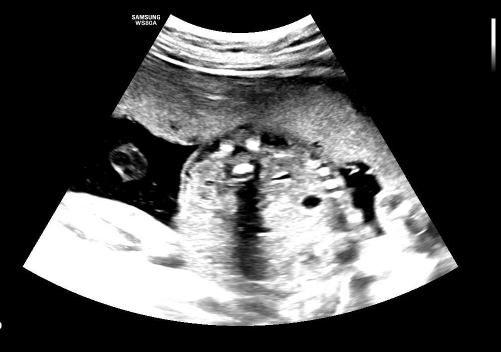}
    \end{subfigure}
    \begin{subfigure}[b]{0.18\textwidth}
        \centering
        \includegraphics[width=\linewidth]{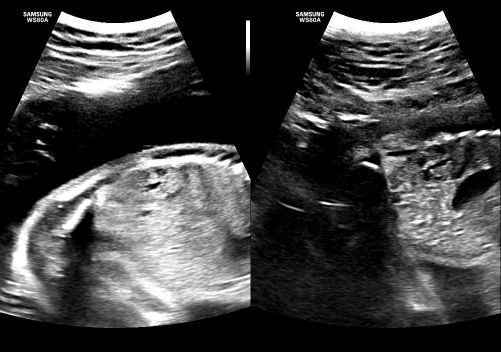}
    \end{subfigure}
    \begin{subfigure}[b]{0.18\textwidth}
        \centering
        \includegraphics[width=\linewidth]{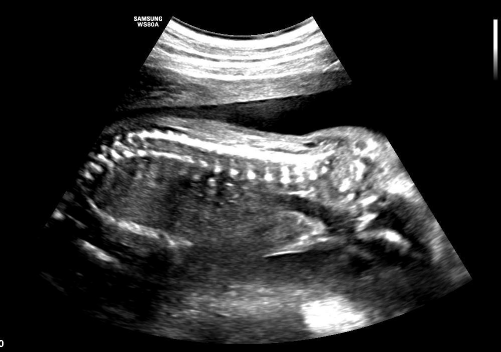}
    \end{subfigure}
    \begin{subfigure}[b]{0.18\textwidth}
        \centering
        \includegraphics[width=\linewidth]{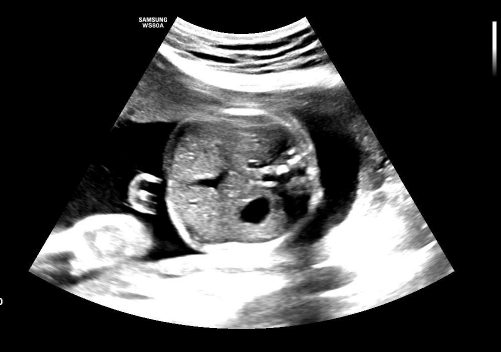}
    \end{subfigure}
    \begin{subfigure}[b]{0.08\textwidth}
        \centering
        {\textbf{C=2.5}}
    \end{subfigure}

    % 第五行：标签列（可选，作为说明文字）
    \vspace{0.5em}
    \caption{Visualization of fetal trunk standard plane images under varying contrast levels (C). Each row represents a specific contrast condition (C = 1.0, 1.5, 2.0, 2.5), and each column corresponds to one of the four target classes: Class 0 (Transverse Kidney), Class 1 (Sagittal Kidney), Class 2 (Sagittal Spine), and Class 3 (Abdominal Circumference). }
    \label{contrast}
\end{figure}

% \begin{figure}[htbp]  
%     \centering  
%     \includegraphics[width=1\linewidth]{Figure/figure2.png} % 假设这个图像包含了两种模型的示意  
%     \caption{Visualization of fetal trunk standard plane images under varying contrast levels (C). Each row represents a specific contrast condition (C = 1.0, 1.5, 2.0, 2.5), and each column corresponds to one of the four target classes: Class 0 (Transverse Kidney), Class 1 (Sagittal Kidney), Class 2 (Sagittal Spine), and Class 3 (Abdominal Circumference). }  
%     \label{contrast}  
% \end{figure}

To better illustrate the impact of varying contrast levels on fetal trunk standard plane images, we present a visual comparison in Figure \ref{contrast}. The figure displays representative images from four anatomical classes (Class 0-3) under four different contrast settings ($C=1.0$, $1.5$, $2.0$, and $2.5$). As the contrast level increases, the local texture and edge features of anatomical structures become progressively more prominent and distinguishable, particularly in soft tissue boundaries and surrounding regions. In contrast, lower contrast conditions tend to preserve overall structural coherence while suppressing local intensity variations.

\subsection{Multi-Contrast Fusion Module}
\begin{figure}[htbp]  
    \centering  
    \includegraphics[width=1\linewidth]{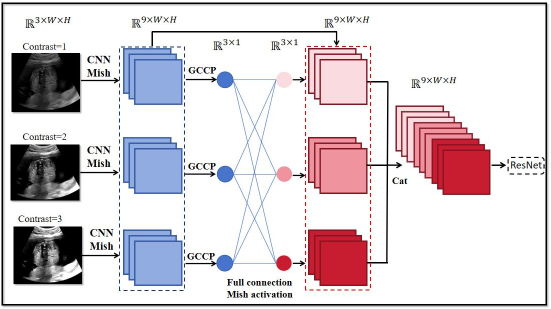} % 假设这个图像包含了两种模型的示意  
    \caption{The whole structure of our Multi-Contrast Fusion Module}  
    \label{model}  
\end{figure}

Our model architecture is illustrated in Figure \ref{model}. The input fetal ultrasound image is first transformed into three contrast-enhanced image sequences, corresponding to contrast levels 1, 2, and 3. The number and specific values of these contrast levels can be flexibly adjusted based on practical requirements.
As shown in the figure, increasing the contrast level from 1 to 3 progressively enhances the visibility of fetal anatomical structures. In low-contrast images (contrast level 1), organ boundaries appear blurry, and only the general shape and outline can be discerned. In contrast, high-contrast images (contrast level 3) clearly reveal fine-grained details such as vascular textures and tissue interfaces—features that are essential for accurate medical diagnosis.
Each set of contrast-enhanced images is independently processed by a feature extraction module composed of convolutional layers followed by the Mish activation function. Each contrast-enhanced image is thus mapped into three feature maps. Compared with the traditional ReLU activation, Mish preserves smooth gradient flow in the negative domain and avoids early saturation in the positive domain due to its unique nonlinear formulation. This design helps mitigate gradient vanishing, which is critical for capturing subtle anatomical features.
In the feature compression stage, we apply our proposed GCCP strategy. For each contrast level, the three extracted feature maps are compressed into a single representative variable, which is then fed into a neural network to learn attention weights. The general formulation of this strategy is as follows:

\begin{equation}
z = \frac{1}{C \times H \times W} \sum_{p=1}^{C} \sum_{i=1}^{H} \sum_{j=1}^{W} I_{pij}
\end{equation}

Where $\mathbf{I}$ denotes the set of all feature maps extracted by the convolutional layers, and $\mathbf{z}$ represents the compressed statistics. The value of $\mathbf{z}$ characterizes the information captured in the feature maps under different contrast settings.
This pooling strategy integrates the complete set of feature maps extracted under each contrast level, enabling the construction of more robust feature representations for both low- and high-contrast images. As a result, it helps to mitigate performance fluctuations caused by variations in image quality.
These statistics $z$ are subsequently fed into a neural network for learning, and its output is passed through a Sigmoid activation function to generate attention weights. 
The attention weights are then assigned to the three feature maps corresponding to each contrast-enhanced image, treating them as a unified group during weighting. After all feature maps are weighted, they are concatenated and the fused representation is fed into a conventional classification backbone, such as ResNet. Therefore, our approach effectively integrates information from multiple contrast levels and can be regarded as a feature fusion technique.

\subsection{Implementation Details}
Experiments were implemented in Python 3.12 and PyTorch 2.4 (CUDA 12.1) on a workstation with Xeon Platinum 8352V CPU and NVIDIA RTX 4090 GPU. Hyperparameters for both the proposed and baselines are detailed in Table \ref{hyperparameters}. We partitioned the dataset into training and test sets using a 70:30 split.
The proposed MCFM module employs four contrast levels [1, 1.3, 1.6, 2] for image enhancement. Convolution operations on each channel extracted three feature maps per adjusted image, generating twelve image representations across all contrast settings. 
Feature maps from each triplet of channels were then averaged, yielding four attention weights corresponding to each contrast-enhanced image version.
We optimized the model using cross-entropy loss as the objective function for classification. Cross-entropy loss, widely used in multi-class classification, quantifies the divergence between predicted probability distributions and ground truth labels, formulated as:
\begin{equation}
\mathcal{L}(x,y)=-\log\left(\frac{e^{x_y}}{\sum_je^{x_j}}\right)=-x_y+\log\left(\sum_je^{x_j}\right)
\end{equation}
Here, x represents the logits (i.e., unnormalized score vector) output by the model for a given input sample, y is the truth label, and $x_{y}$ refers to the logit associated with the correct class. 
The function first applies a softmax transformation to convert the model output into a probability distribution, and then calculates the negative log-likelihood between the predicted and actual distributions. 
A lower loss value indicates that the model’s prediction is more accurate and closer to the true label.

\begin{table}[ht]
\centering
\caption{Training hyperparameters used in the experiments}
\begin{tabular}{p{5cm}p{3cm}}
\hline
\textbf{Hyperparameter} & \textbf{Value} \\
\hline
Batch size & 64 \\
Epoch & 20 \\
Lr & 0.001 \\
Optimizer & Adam \\
Loss Function & CrossEntropy \\
\hline
\end{tabular}
\label{hyperparameters}
\end{table}

\section{Result}
\subsection{Evaluation Metrics}

To comprehensively assess the performance of our model, we adopt several standard evaluation metrics, including Precision, Recall, F1-score, the Receiver Operating Characteristic (ROC) curve, the Precision-Recall (PR) curve, the confusion matrix, the Area Under the ROC Curve (AUC), and the Average Precision (AP). Precision and Recall are fundamental metrics for evaluating classification tasks, particularly in the presence of class imbalance. Precision is defined as the proportion of true positive predictions among all predicted positives, formulated as
\begin{equation}
\text{Precision} = \frac{TP}{TP + FP},
\end{equation}
where $TP$ and $FP$ denote the numbers of true positives and false positives, respectively. Recall measures the proportion of true positive predictions among all actual positive instances, given by
\begin{equation}
\text{Recall} = \frac{TP}{TP + FN},
\end{equation}
where $FN$ denotes the number of false negatives. The F1-score provides a harmonic mean of Precision and Recall, and is expressed as
\begin{equation}
\text{F1} = 2 \times \frac{\text{Precision} \times \text{Recall}}{\text{Precision} + \text{Recall}}.
\end{equation}
In addition, the confusion matrix presents a complete summary of classification results by reporting the counts of true positives (TP), false positives (FP), true negatives (TN), and false negatives (FN). The ROC curve illustrates the trade-off between the true positive rate and the false positive rate at various thresholds, and its corresponding AUC quantifies the overall ability of the model to distinguish between classes. Similarly, the PR curve focuses on the relationship between precision and recall across different thresholds, which is especially informative in imbalanced settings. The Average Precision (AP) summarizes the PR curve by calculating the area under it, reflecting the average model precision over all levels of recall.

\subsection{Baseline models}
To comprehensively evaluate the proposed method, five representative image classification models were selected as baselines, encompassing various network architectures, including lightweight convolutional networks, deep feature extractors, and Transformer-based vision models. 
The selected models include EfficientNet\cite{tan2019efficientnet}, Inception V3\cite{szegedy2016rethinking}, VGG11\cite{simonyan2014very}, Vision Transformer (ViT)\cite{Dosovitskiy2020AnII}, and Swin Transformer\cite{liu2021swin}.

EfficientNet, a lightweight model, achieves competitive performance with relatively low computational costs, making it well-suited for classification tasks in resource-limited environments. 
Inception V3 utilizes multi-scale convolutional kernels to expand receptive fields and enrich feature representations. 
It also incorporates techniques such as factorization and auxiliary classifiers, leading to strong feature extraction capabilities. 
VGG11, a classic CNN architecture, features a simple design and moderate depth. 
It builds deep models by stacking multiple small convolutional kernels and has demonstrated strong performance in various image recognition tasks. 
ViT was the first to introduce Transformer architectures into visual tasks, enabling global context modeling by dividing images into patches and flattening them into token sequences. 
To enhance Transformer performance in visual tasks, Swin Transformer introduced a hierarchical architecture and a sliding window attention mechanism. 
These improvements increased local modeling efficiency while preserving global context awareness, establishing it as a leading vision Transformer model in recent years.

\subsection{Comparison Experiments}
Table \ref{baselines} summarizes the classification performance of various models for this task. 
As shown in the results, MCFM-ResNet18 achieved the highest performance across all evaluation metrics, with an accuracy of 95.83\% and an F1-score of 95.92\%, significantly outperforming other models. 
MCFM-ResNet34 also exhibited stable performance, achieving an F1-score of 95.20\%. 
These results highlight the strong adaptability and generalization capability of the MCFM module across different backbone architectures.
In contrast, conventional convolutional models—such as VGG11, Inception V3, and EfficientNet---exhibited similar performance, with all F1-scores below 91\%. 
This indicates their limited ability to capture task-specific discriminative features. By comparison, the MCFM-based models more effectively extracted salient features from the images.
Notably, Transformer-based models, including ViT and Swin Transformer, performed poorly on this task. 
ViT achieved an accuracy of only 34.44\%, while Swin Transformer performed even worse, at just 25.83\%. 
Both models also showed a marked decline in metrics such as Precision and F1-score. 
This may be attributed to the inherent reliance of Transformer architectures on large-scale datasets and long-range dependency modeling. 
Due to the limited dataset size and the importance of local feature extraction in this task, the architectural strengths of Transformers could not be fully utilized.

\begin{table}[ht]
\centering
\caption{Performance comparison across various baseline models}
\begin{tabular}{p{3cm}p{2cm}p{2cm}p{2cm}p{2cm}p{1.7cm}}
\hline
\textbf{Model} & \textbf{ACC} & \textbf{Precision} & \textbf{Recall} & \textbf{F1-score} & \textbf{Parameter} \\
\hline
MCFM-ResNet18 & 0.9583 & 0.9604 & 0.9590 & 0.9592 & 11.21M \\
MCFM-ResNet34 & 0.9528 & 0.9554 & 0.9503 & 0.9520 & 21.32M \\
EfficientNet & 0.8889 & 0.8983 & 0.8866 & 0.8840 & 4.01M \\
Inception V3 & 0.8944 & 0.9044 & 0.8936 & 0.8933 & 24.35M \\
VGG11 & 0.9083 & 0.9098 & 0.9090 & 0.9092 & 128.78M \\
Vit & 0.3444 & 0.4220 & 0.3351 & 0.3055 & 85.80M \\
Swin Transformer & 0.2583 & 0.0646 & 0.2500 & 0.1026 & 27.52M \\
\hline
\end{tabular}
\label{baselines}
\end{table}

\subsection{Ablation Experiment}
To further validate the effectiveness of the proposed MCFM module, ablation studies were performed by integrating it into two widely used backbone architectures—ResNet18 and ResNet34—and comparing the modified networks with their original versions. The experimental results are summarized in Table \ref{backbones}.

\begin{table}[ht]
\centering
\caption{Performance comparison details of our module implemented in ResNet series models}
\begin{tabular}{p{3cm}p{2cm}p{2cm}p{2cm}p{2cm}p{1.7cm}}
\hline
\textbf{Model} & \textbf{ACC} & \textbf{Precision} & \textbf{Recall} & \textbf{F1-score} & \textbf{Parameter} \\
\hline
ResNet18 & 0.9111 & 0.9171 & 0.9121 & 0.9119 & 11.18M \\
MCFM-ResNet18 & 0.9583 & 0.9604 & 0.9590 & 0.9592 & 11.21M \\
ResNet34 & 0.9222 & 0.9258 & 0.9238 & 0.9231 & 21.29M \\
MCFM-ResNet34 & 0.9528 & 0.9554 & 0.9503 & 0.9520 & 21.32M \\
\hline
\end{tabular}
\label{backbones}
\end{table}

As shown in the results, the inclusion of the MCFM module significantly improved performance across both backbone networks. 
For example, the original ResNet18 achieved 91.11\% accuracy, 91.71\% precision, 91.21\% recall, and 91.19\% F1-score. By contrast, MCFM-ResNet18 attained 95.83\% accuracy, 96.04\% precision, 95.90\% recall, and 95.92\% F1-score. 
Notably, the F1-score increased by 4.7 percentage points, highlighting the module’s effectiveness in enhancing discriminative capability.
Similarly, the MCFM module also improved the overall performance of the deeper ResNet34 backbone. 
The original ResNet34 achieved 92.22\% accuracy and 92.31\% F1-score, while MCFM-ResNet34 achieved 95.28\% and 95.20\%, respectively. 
These results highlight the MCFM module’s strong compatibility and enhancing effect across backbones of varying depth.
Although the MCFM module significantly improved performance, its impact on model complexity should also be considered. 
Specifically, integrating the MCFM module into ResNet18 and ResNet34 added approximately 0.3 million parameters to each model. 
Despite the increase in model size, the substantial improvements in ACC, Precision, Recall, and F1-score indicate that the added complexity is acceptable for practical applications.

\begin{figure}[htbp]  
    \centering  
    \includegraphics[width=1\linewidth]{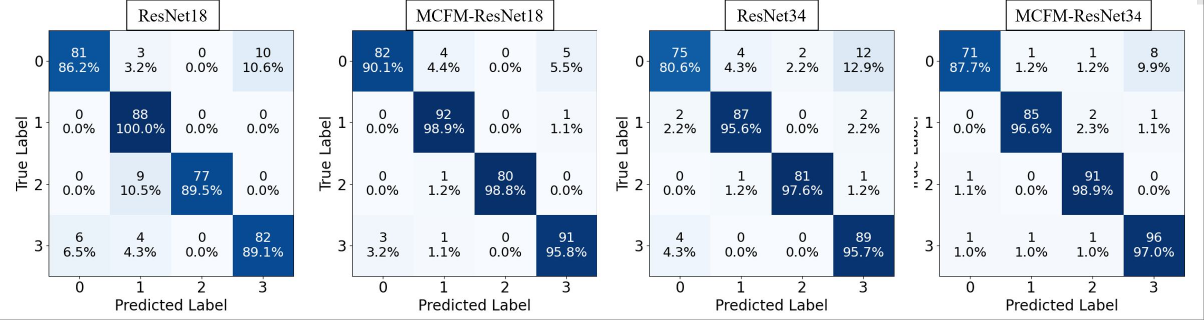} 
    \caption{Confusion matrices of ResNet18, MCFM-ResNet18, ResNet34, and MCFM-ResNet34 on the fetal standard plane classification task. The introduction of the Multi-Context Fusion Module (MCFM) consistently improves classification performance across all categories, demonstrating its effectiveness in enhancing feature representation and class discrimination.}  
    \label{conf}  
\end{figure}

Figure \ref{conf} shows the confusion matrices of different models in the four-class classification task, further validating the MCFM module’s effectiveness in improving discriminative performance. 
Compared with the original ResNet18, MCFM-ResNet18 significantly improved classification accuracy for Class 0 and Class 3, with fewer incorrect predictions. 
Specifically, accuracy for Class 0 rose from 86.2\% to 90.1\%, and for Class 3 from 89.1\% to 95.8\%.
Similarly, MCFM integration into the deeper ResNet34 network resulted in consistent improvements. 
MCFM-ResNet34 showed improved accuracy across all classes, particularly in Class 3, where performance increased from 95.7\% to 97.0\%. 
It also reduced inter-class misclassification, further demonstrating MCFM’s effectiveness in enhancing both generalization and fine-grained discrimination. 
Overall, the MCFM module significantly improved the model’s ability to distinguish challenging samples and reduced inter-class confusion.

\begin{figure}[htbp]  
    \centering  
    \includegraphics[width=1\linewidth]{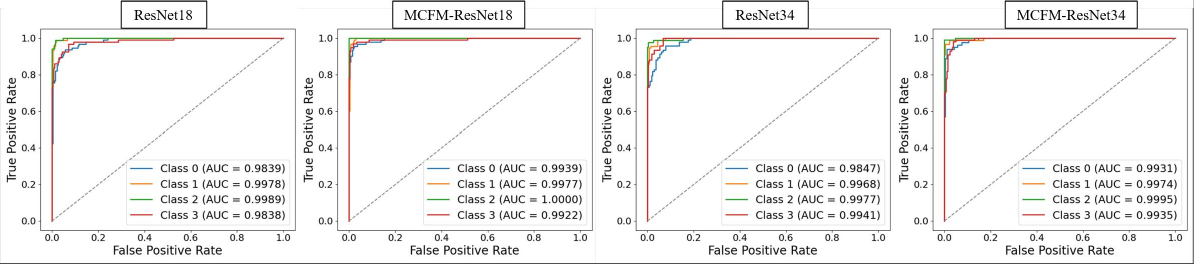} 
    \caption{ROC curves and AUC values of ResNet18, MCFM-ResNet18, ResNet34, and MCFM-ResNet34 across four standard plane classes. The MCFM-enhanced models demonstrate consistently higher AUC scores, indicating improved classification confidence and discriminative power compared to the baseline architectures.}  
    \label{roc}  
\end{figure}

Figure \ref{roc} illustrates the ROC curves and corresponding AUC values for four models—ResNet18, MCFM-ResNet18, ResNet34, and MCFM-ResNet34—on the four-class classification task. With the integration of MCFM, AUC scores for all classes improved. Notably, MCFM-ResNet18 achieved a perfect AUC of 1.0000 for Class 2, while all four class AUCs for MCFM-ResNet34 exceeded 0.993, indicating enhanced class boundary recognition. These results suggest that incorporating the MCFM module not only improves overall accuracy but also enhances fine-grained class discrimination.

\begin{figure}[htbp]  
    \centering  
    \includegraphics[width=1\linewidth]{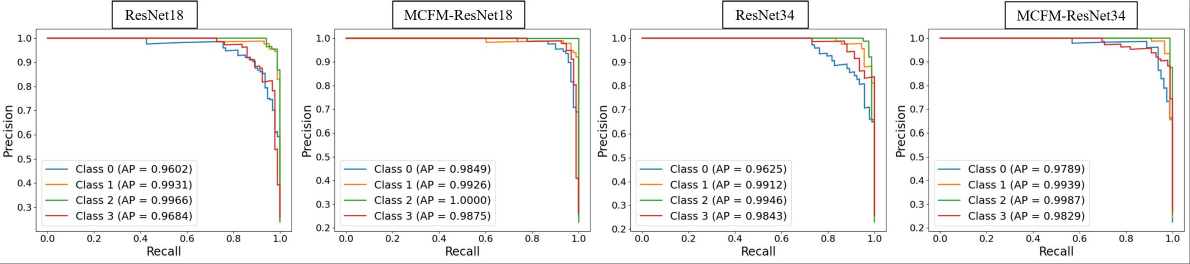} 
    \caption{Precision–Recall (PR) curves and Average Precision (AP) scores of ResNet18, MCFM-ResNet18, ResNet34, and MCFM-ResNet34 on the standard plane classification task. The proposed MCFM module consistently improves AP values across all classes, highlighting its robustness in handling class imbalance and enhancing prediction reliability.}  
    \label{pr}  
\end{figure}

Figure \ref{pr} presents the PR curves of the four models in the four-class classification task, further illustrating the trade-off between precision and recall across different classes. 
The PR curves for MCFM-ResNet18 and MCFM-ResNet34 lie closer to the upper right corner compared to their original counterparts, indicating more robust predictive performance with reduced false positives and false negatives across all classes.
In contrast, the original ResNet18 and ResNet34 exhibit notable fluctuations in certain classes, with unsmooth PR curves and reduced precision at high recall levels, reflecting less stable performance and weaker discrimination in high-recall scenarios. 
In contrast, the MCFM-enhanced models demonstrate smoother and more consistent PR curves across all classes, indicating improved robustness and superior overall discriminative capability in multi-class classification tasks.

\section{Discussion}  

\subsection{Clinical Significance}
Pregnancy prenatal ultrasound plays a critical role in the early detection of fetal growth abnormalities, and accurate identification of structural anomalies is essential for reducing perinatal complications and neonatal mortality. This study focuses on the automated classification of fetal trunk planes—including transverse kidney, sagittal kidney, sagittal spine, and transverse abdomen—offering reliable support for the accurate screening and management of fetal growth disorders.  As the structural core of the fetal body, the trunk plays a vital role in maintaining overall fetal health. The kidneys, as essential excretory organs, are closely associated with fetal metabolic regulation and homeostasis. Identifying the transverse and sagittal kidney planes enables early detection of renal dysplasia, hydronephrosis, polycystic kidney disease, and other renal anomalies. The fetal spine, as the central component of the axial skeleton, provides structural support and enables motor function. The sagittal view of the spine allows for the assessment of spinal continuity, physiological curvature, and vertebral morphology, facilitating the diagnosis of conditions such as spina bifida and tethered cord syndrome. The transverse abdomen view reveals the position, size, and morphology of abdominal organs such as the liver and spleen, which are critical for evaluating fetal abdominal development. The proposed model achieved an accuracy of approximately 95\%, clearly demonstrating its effectiveness. Before physician review, the model can generate preliminary predictions for each fetal plane, providing corresponding classification labels. Physicians can then perform a secondary screening based on these labels, improving diagnostic efficiency and reducing cognitive workload.

\begin{table}[ht]
\centering
\caption{Classification performance of MCFM-ResNet in different classes}
\begin{tabular}{p{3cm}p{3cm}p{3cm}p{3cm}}
\hline
\textbf{Class} & \textbf{Precision} & \textbf{Recall} & \textbf{F1-score} \\
\hline
Class 0 & 0.9647 & 0.9011 & 0.9318 \\
Class 1 & 0.9388 & 0.9892 & 0.9633 \\
Class 2 & 1.0000 & 0.9877 & 0.9938 \\
Class 3 & 0.9381 & 0.9579 & 0.9479 \\
\hline
\end{tabular}
\label{differentclass}
\end{table}

The MCFM-ResNet18 model demonstrates robust performance across all categories, as shown in Table \ref{differentclass}. The sagittal spine view achieved the highest accuracy (F1-score of 0.99), followed by the sagittal kidney view and transverse abdominal plane, with F1-scores of 0.96 and 0.95, respectively. The transverse kidney view performed slightly lower (F1-score of 0.93), but the model still demonstrated strong recognition capabilities, highlighting its stability and practical utility.

\subsection{Differences to Previous Models}
Considering the inherent limitations of ultrasound images, especially their deficiencies in expressing detailed information, we designed the Multi-Contrast Fusion Module (MCFM) to enhance the model's ability to model detailed features. Unlike traditional methods such as SENet \cite{hu2018squeeze} and FcaNet \cite{qin2021fcanet}, the proposed attention mechanism is specifically applied to the lower layers of the model, directly acting on the raw images that have not been convoluted. We believe that SENet and FcaNet find it difficult to effectively model and assign weights to raw images with different contrasts. SENet relies on global average pooling to compress and extract statistical features from images. However, raw images are usually large in size and have not yet been convoluted to extract effective features, containing a lot of noise and redundant information. Therefore, direct compression may not accurately reflect the essential features of the image. FcaNet introduces frequency-domain information for modeling, which can filter out some interference and enhance the retention of effective information. But at the level of raw images, it still finds it hard to overcome the redundancy and interference in images with different contrasts, thus affecting performance. Our research addresses this by using meta-convolution for preliminary feature extraction of images with different contrasts. Then, the GCCP strategy is adopted to perform overall compression and averaging operations on all feature maps extracted from contrast images through convolution. The resulting statistics are more representative and can be used with meta-convolution to filter out redundancy and interference. Finally, GCCP is used for one-time compression to obtain a better representation.

\subsection{Advantages, Drawbacks and Future Works}
The attention module is exclusively integrated into the lower layers and directly applied to planes with varying contrast levels. 
As a result, the module adds minimal computational overhead. Compared to conventional methods, the proposed module improves the extraction of fine-grained features and global context, leading to enhanced overall performance. 
However, the model has some limitations. Specifically, the employed compression strategy averages all feature maps under a single contrast setting into one aggregated value. 
This may result in significant information loss. Furthermore, the relatively small dataset size may hinder the model’s ability to learn robust and generalizable representations. 
Future work will explore more effective compression techniques, potentially leveraging frequency-domain approaches. 
Additionally, we plan to construct a larger dataset to enable learning from a broader and more diverse set of samples.

\section{Conclusion}
This study introduces a novel deep learning framework, MCFM, to address the challenges of inconspicuous details and blurred key information in the automatic recognition of fetal trunk plane. Our model is only integrated into the lower layers of the model, assigning weights to the input sections under different contrasts. We map the input trunk section into copies under different contrasts, and then introduce a meta-convolution to each copy to filter out valid information. Then we introduce the GCCP strategy to compress all the feature maps extracted from an image under one contrast at one time, obtaining a statistical representation. After that, it is input into the neural network for learning to obtain attention weights. We applied our model to our dataset and conducted comparative and ablation experiments. The experimental results show that after incorporating the MCFM module, the model's parameters did not increase significantly, but the model's performance was greatly enhanced.

\bibliographystyle{ieeetr} % 选择参考文献的格式，如 plain, unsrt, alpha, ieeetr 等  
\bibliography{reference.bib} % 指定.bib文件的名称（无需文件扩展名）

\section*{Figure Legends}
\begin{itemize}

    \item Figure 1: Representative samples from each class in the dataset. The four classes include: Class 0 – Transverse Kidney, Class 1 – Sagittal Kidney, Class 2 – Sagittal Spine, and Class 3 – Transverse Abdomen. 

    \item Figure 2: Visualization of fetal trunk standard plane images under varying contrast levels (C). Each row represents a specific contrast condition (C = 1.0, 1.5, 2.0, 2.5), and each column corresponds to one of the four target classes: Class 0 (Transverse Kidney), Class 1 (Sagittal Kidney), Class 2 (Sagittal Spine), and Class 3 (Abdominal Circumference).

    \item Figure 3: The whole structure of our Multi-Contrast Fusion Module

    \item Figure 4: Confusion matrices of ResNet18, MCFM-ResNet18, ResNet34, and MCFM-ResNet34 on the fetal standard plane classification task. The introduction of the Multi-Context Fusion Module (MCFM) consistently improves classification performance across all categories, demonstrating its effectiveness in enhancing feature representation and class discrimination. 

    \item Figure 5: ROC curves and AUC values of ResNet18, MCFM-ResNet18, ResNet34, and MCFM-ResNet34 across four standard plane classes. The MCFM-enhanced models demonstrate consistently higher AUC scores, indicating improved classification confidence and discriminative power compared to the baseline architectures.

    \item Figure 6: Precision–Recall (PR) curves and Average Precision (AP) scores of ResNet18, MCFM-ResNet18, ResNet34, and MCFM-ResNet34 on the standard plane classification task. The proposed MCFM module consistently improves AP values across all classes, highlighting its robustness in handling class imbalance and enhancing prediction reliability.
\end{itemize}

\end{document}